\documentclass[10pt,twocolumn,letterpaper]{article}

\usepackage{cvpr}              

\newcommand{\bx}{\mathbf{x}}
\newcommand{\bl}{\mathbf{l}}
\newcommand{\bv}{\mathbf{v}}
\newcommand{\bu}{\mathbf{u}}
\newcommand{\bp}{\mathbf{p}}
\newcommand{\ba}{\mathbf{a}}
\newcommand{\bs}{\mathbf{s}}
\newcommand{\bt}{\mathbf{t}}
\newcommand{\bbR}{\mathbbm{R}}

\definecolor{cvprblue}{rgb}{0.21,0.49,0.74}
\usepackage[pagebackref,breaklinks,colorlinks,allcolors=cvprblue]{hyperref}
\usepackage{bbm}
\usepackage{multirow}
\usepackage{caption}
\usepackage{subcaption}

\def\paperID{*****} 
\def\confName{CVPR}
\def\confYear{2025}

\title{Understanding the Effect of using Semantically Meaningful Tokens for Visual Representation Learning}

\author{Neha Kalibhat\textsuperscript{1} \\
  {\tt\small nehamk@cs.umd.edu} \\
  \and
  Priyatham Kattakinda\textsuperscript{1} \\
  \and
  Sumit Nawathe\textsuperscript{1} \\
  \and
  Arman Zarei\textsuperscript{1} \\
  \and
  Nikita Seleznev\textsuperscript{2} \\
  \and
  Samuel Sharpe\textsuperscript{2} \\
  \and
  Senthil Kumar\textsuperscript{2} \\
  \and
  Soheil Feizi\textsuperscript{1} \\
  \and
  \textsuperscript{1}University of Maryland
  \and
  \textsuperscript{2}CapitalOne Research
}

\begin{document}
\maketitle


\begin{abstract}
Vision transformers have established a precedent of patchifying images into uniformly-sized chunks before processing. We hypothesize that this design choice may limit models in learning comprehensive and compositional representations from visual data. This paper explores the notion of providing semantically-meaningful visual tokens to transformer encoders within a vision-language pre-training framework. Leveraging off-the-shelf segmentation and scene-graph models, we extract representations of instance segmentation masks (referred to as tangible tokens) and relationships and actions (referred to as intangible tokens). Subsequently, we pre-train a vision-side transformer by incorporating these newly extracted tokens and aligning the resultant embeddings with caption embeddings from a text-side encoder. To capture the structural and semantic relationships among visual tokens, we introduce additive attention weights, which are used to compute self-attention scores. Our experiments on COCO demonstrate notable improvements over ViTs in learned representation quality across text-to-image ($+47\%$) and image-to-text retrieval ($+44\%$) tasks. Furthermore, we showcase the advantages on compositionality benchmarks such as ARO ($+18\%$) and Winoground ($+10\%$).
\end{abstract}
\begin{figure*}
    \centering
    \includegraphics[width=\textwidth]{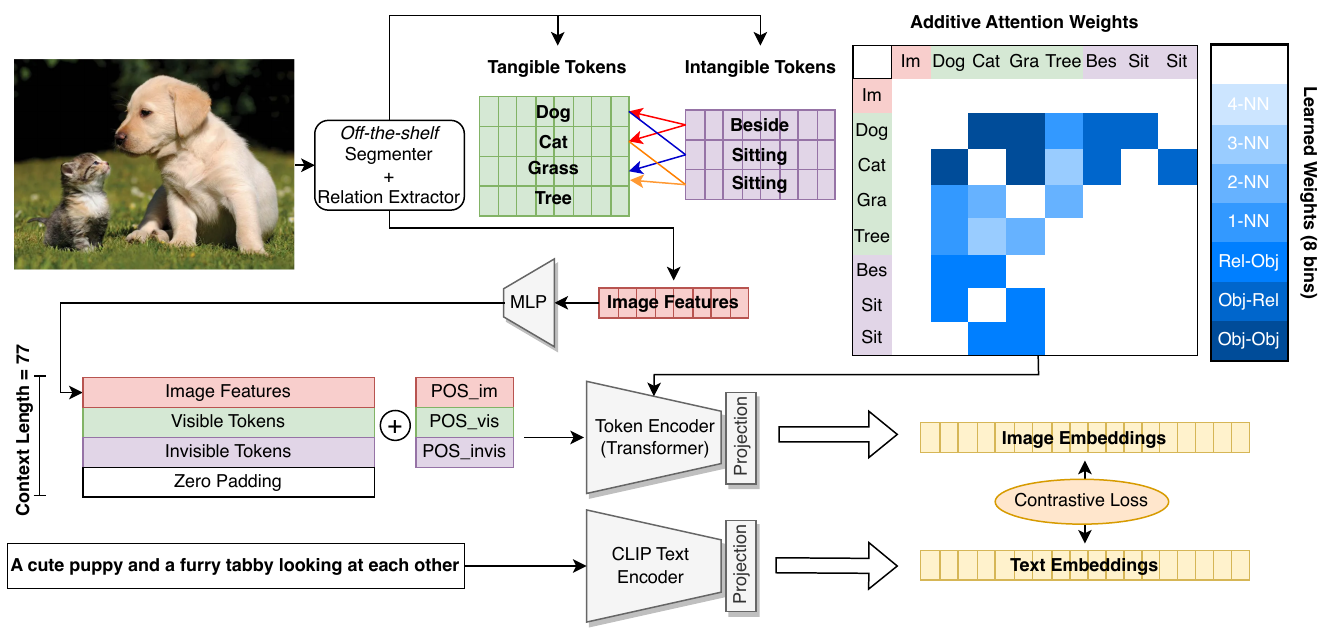}
    \caption{\textbf{Training with meaningful visual tokens:} We present a framework that uses off-the-shelf segmentation and relation extraction models to prepare a set of \textit{tangible tokens} ($\mathcal{V}$) and \textit{intangible tokens} ($\mathcal{U}$) for any arbitrary image, along with directional semantic relationships between them. These tokens and image features ($\bl$) are then passed as input to our visual token encoder ($f(.)$). We utilize the semantic relation ($\mathcal{E}$) and relative location ($\mathcal{N}$) information of all tokens to compute additive attention weights, ranked by importance. The learned image embeddings ($\bs$) are contrastively aligned with the text embeddings ($\bt$) of the CLIP text encoder ($g(.)$), which is simultaneously fine-tuned with our model.}
    \label{fig:framework}
\end{figure*}
\section{Introduction}
Vision transformers (ViTs) \cite{dosovitskiy2021image, liu2021swin, touvron2020training} have emerged as a groundbreaking innovation in the field of computer vision, leveraging the power of transformer architectures \cite{vaswani2023attention} to process and interpret visual data. They also have shown unprecendented performance in multi-modal setups \cite{radford2021learning, Singh_2022, zeng2022multigrained, li2019visualbert, li2022blip} where vision and language data are aligned for downstream retrieval, question-answering, captioning and other tasks. Tokenization is a key feature of transformers where the input is split into small chunks which are converted into vectors and then processed by the model. Text sentences (in the English language) is generally tokenized into words. Image data for ViTs is generally split into a grid of equally sized patches and then flattened into a sequence. Although this practice is widely accepted, we propose a different approach to patchifying which attempts to capture more high-level semantic information in patches. 

Using off-the-shelf segmentation and scene-graph generation techniques, we extract panoptic segmentation mask embeddings of all objects a given image and CLIP \cite{radford2021learning} text embeddings of the actions or relationships between them. We call the object embeddings as \textit{tangible tokens} since these are visible in the image and the relationship embeddings as the \textit{intangible tokens} since these are not visible but are still observable. Both sets of tokens have independent semantic meaning, similar to words in a sentence but unlike equal-sized patches in ViTs. We also extract other metadata including image features, [subject, object, predicate] triplets and K-nearest neighbors in the image for each object. This metadata helps us capture both directional relationship and relative position information for complete visual comprehension. 

We demonstrate a proof-of-concept that applies such semantically-meaningful tokens in visual representation learning and study its potential. We train a transformer model (called Visual Token Encoder) on the set extracted tokens of the COCO \cite{lin2015microsoft} dataset. We apply an additive attention mechanism using the relational and structural information from the metadata, ranked by importance. The learned image embeddings are contrastively aligned with the COCO caption embeddings from the CLIP text encoder which is fine-tuned alongside our model. We compare our method with $2$ other vision-language pre-training setups which follow the same training regime as ours but the image-side encoder is replaced with (i) A ViT (randomly-initialized) or (ii) The CLIP image encoder (fine-tuned) and trained directly on COCO images. 

Our experiments show that our tokenization process significantly improves representation quality, resulting in a $47\%$ improvement in text-to-image retrieval over a ViT and $9\%$ over CLIP (fine-tuned) on the COCO validation split. Moreover, we show improved compositional reasoning capabilities of the learned image representations by exploring the ARO \cite{yuksekgonul2023visionlanguage} and Winoground \cite{Thrush_2022} benchmarks. Our Visual Token Encoder outperforms the ViT by $18\%$ on ARO and $9\%$ on Winoground. These results indicate a promising angle for upcoming research - to re-think better encoder architectures that encapsulate high-level, semantic entities for improved visual understanding.
\begin{figure*}
    \centering
    \includegraphics[width=\textwidth]{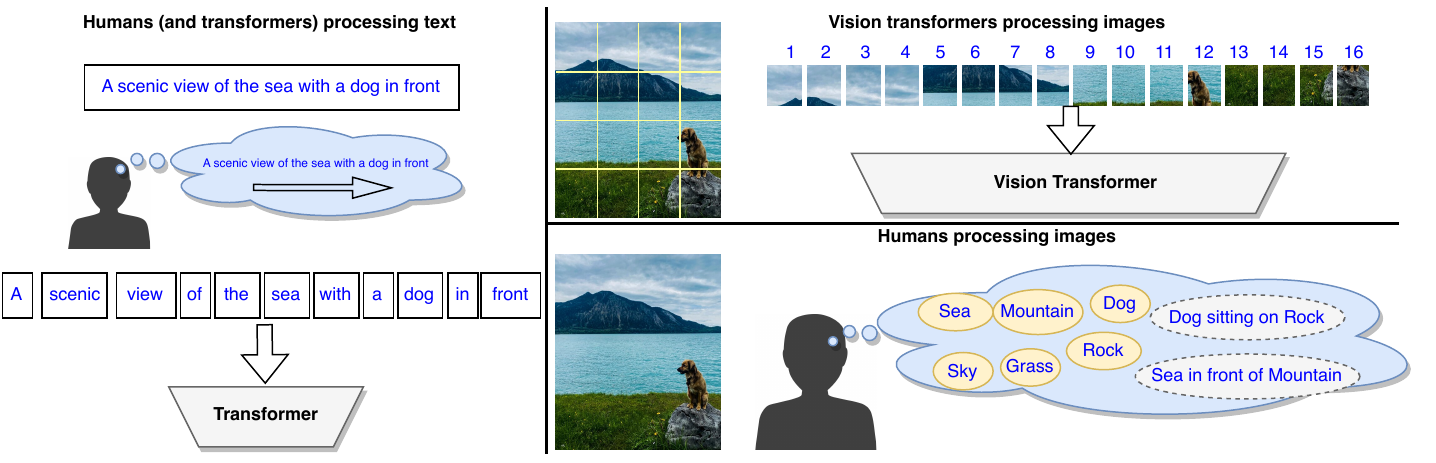}
    \caption{\textbf{Processing of text vs image data:} In this simple illustration, we demonstrate the notable difference in how text and visual data are processed by humans and transformers. Humans are capable of deciphering larger concepts from images (both tangible and intangible), where each concept has independent semantic meaning.}
    \label{fig:visual_tokens}
\end{figure*}
\section{Related Work}

\paragraph{Transformers and ViTs.} Transformers \cite{vaswani2023attention} have revolutionized the field of natural language processing and have been increasingly applied to computer vision tasks. The self-attention mechanism in transformers enables the model to capture long-range dependencies, making them highly effective for various applications. Vision Transformers (ViTs) \cite{dosovitskiy2021image} introduced the concept of patching, where an image is divided into fixed-size patches, and these patches are treated as tokens similar to words in NLP tasks. This patching approach allows transformers to process images efficiently, leveraging their powerful attention mechanisms. However, ViTs often struggle with computational efficiency and local feature extraction. Subsequent ViT variants, such as Swin Transformers \cite{liu2021swin}, have introduced hierarchical structures that enhance the model's ability to capture multi-scale features. These hierarchical transformers divide the image into non-overlapping windows and compute self-attention within each window, allowing for better handling of larger images and more detailed features.

\paragraph{Large-Scale Image Segmentation.} In the realm of large-scale image segmentation, several prominent models have made significant strides in addressing complex visual tasks. Among these, the Segment Anything Model (SAM) \cite{alex2023segment} and Segment Everything Everywhere All At Once (SEEM) \cite{zou2023segment} stand out due to their innovative approaches. These models excel at dividing an image into different components and parts, enabling detailed analysis and interpretation of complex scenes by segmenting and categorizing each visual element distinctly. Building on their innovative frameworks, they are adept at performing a variety of segmentation tasks, including semantic segmentation \cite{long2015fully}, panoptic segmentation \cite{kirillov2019panoptic}, and instance segmentation \cite{he2018mask}.

\paragraph{Architectural Improvements.} To enhance understanding of complex images and processing them more efficiently, various methods have explored novel techniques. \cite{han2022vision} introduced Vision GNN (ViG), which models images as graphs by treating patches as nodes and their relationships as edges, effectively capturing complex structures and spatial relationships within images, thus outperforming traditional CNNs and transformers on some benchmarks. In another approach, \cite{ma2024groma} presented Groma, a Multimodal Large Language Model (MLLM) utilizing localized visual tokenization to handle region-level tasks effectively, demonstrating superior performance on COCO \cite{lin2015microsoft} and Visual Genome benchmarks by efficiently grounding textual outputs to specific image regions. \cite{xia2023dat} proposed the Deformable Multi-Head Attention (DMHA) module in the Deformable Attention Transformer (DAT), which dynamically allocates key and value pairs to relevant regions, enhancing representation power while reducing computational overhead, achieving state-of-the-art results on benchmarks like ImageNet \cite{imagenet}, MS-COCO \cite{lin2015microsoft}, and ADE20K \cite{zhou2018semantic}. Additionally, \cite{Chen_2021} introduced the Deformable Patch-based Transformer (DPT), featuring a Deformable Patch (DePatch) module that dynamically adjusts patch positions and scales to preserve local structures and semantic integrity, thereby significantly improving performance in image classification and object detection tasks.

\begin{figure*}
    \centering
    \includegraphics[width=\textwidth,trim={0.2cm 0cm 0cm 0.1cm},clip]{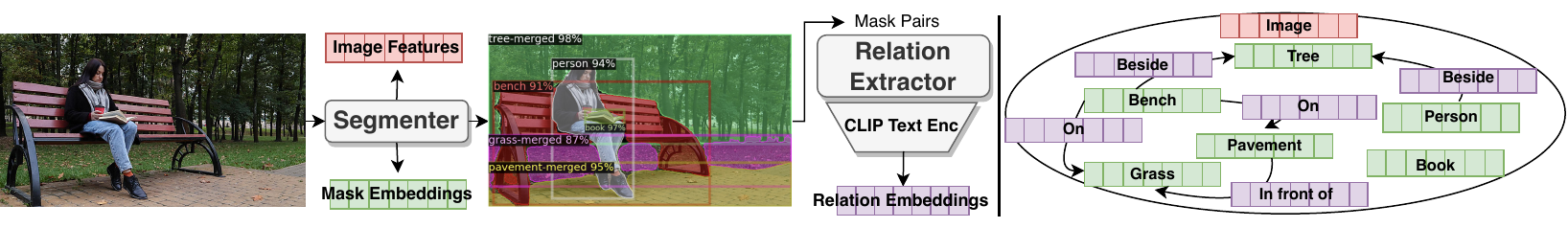}
    \caption{\textbf{Using off-the-shelf models to extract tokens:} We extract image features ($\bl$) and mask embeddings ($\mathcal{V}$) from a panoptic segmentation model. Next, we pass pairs of object masks to a relation extractor and collect the highly probable relationships ($\mathcal{E}$). We compute CLIP text embeddings of all relationships ($\mathcal{U}$). This information is distilled into a scene graph representing the image as shown. }
    \label{fig:seg_rel}
\end{figure*}
\section{Re-thinking Tokenization in Vision Transformers}
\label{sec:tokenization}


\paragraph{Text:} Given a text corpus, tokenizers are typically constructed to concisely represent the text by capturing the sentences with as few tokens as possible while maintaining meaningful information. A simple yet effective practice to tokenize is, by splitting a given sentence into constituent words using delimiting characters like spaces, periods, commas, etc. This results in words that have semantic (sometimes physical) meaning when considered independently as well as in the sentence context. In Figure \ref{fig:visual_tokens}, we show an example of the sentence \textit{A scenic view of the sea with a dog in front} and tokenize it into words such as \textit{a, scenic, view, of, the, sea,} etc. In majority of written languages (including English) humans are conditioned to read a sentence word by word (right to left or left to right depending on the language). As we read each word, we associate that word with its meaning and simultaneously deduce the meaning of the sentence. The design of transformers and tokenization therefore makes intuitive sense as it closely mimics how humans process sentences.

\paragraph{Image:} Visual tokenizers (in ViTs), on the other hand, treat images as a grid of patches, which are then flattened into a sequence and processed with several multi-head attention layers. Specifically, an image of size $224\times224$ is divided into small, equally-sized patches (say $16\times16$), resulting in $196$ patches or \textit{tokens} which are processed as if they were a sequence. This process of tokenization was adapted from transformers for text data. But unlike text data, each visual token (or patch) does not always have independent semantic meaning. This is illustrated in Figure \ref{fig:visual_tokens}, where tokens $3$, $4$, $13$, and $16$, when examined independently, have ambiguous semantic and physical meaning. For example, token $3$ and $4$ can be associated with a blue marble stone, and $16$ can be associated with brushed metal. Each token needs to be studied in the context of the surrounding tokens or the entire image to be associated with a physical meaning. Therefore, there is a fundamental difference in the significance of text tokens and visual tokens although they are processed by transformer architectures in almost identical manners.


An alternative approach is to divide up the image into larger entities that each have independent physical meaning. Each of these entities possesses several observable constituent attributes. For example, in Figure \ref{fig:visual_tokens}, we have \textit{sky, mountain, grass, sea, dog}, etc., where the grass is green with small yellow flowers and the mountain is rocky with alpine trees. Apart from these physical entities, we also draw several conclusions that are not necessarily associated with tangible (or visible) concepts. For example, the sea is in front of the mountain, the dog is sitting on the rock, and the rock lies on the grass. Both sets of entities, tangible and intangible, play a vital role in fully comprehending the image.

Our observation leads us to propose a set of modifications to the transformer architecture so that images are tokenized and processed in a more semantically meaningful way. In the next section, we define tangible and intangible tokens and describe how these can be extracted for any given image using off-the-shelf models. We hypothesize that this tokenization strategy facilitates the transformer's ability to process and reason about the various objects and their interrelationships, as each high-level visual entity is represented as an individual token readily accessible to the transformer.

\section{Approach}
\subsection{Using Off-the-shelf Models to Extract Visual Tokens}\label{sec:seg_rel}
Let's consider a real-world image $\bx$ which may contain several entities, both in the foreground and background. We define the set of tangible entities (tokens) as $\mathcal{V}$. $\mathcal{V}$ may include several items of varying sizes, ranging from small details like the coffee cup in Figure \ref{fig:seg_rel} to much larger entities like the trees in the background. As discussed in Section \ref{sec:tokenization}, there are several observable components in the image that cannot be localized. These entities usually correspond to actions or relationships among objects in the scene. We denote this set of \textit{intangible entities (tokens)} as $\mathcal{U}$.

We utilize an off-the-shelf instance segmentation model titled Segment Everything Everywhere All At Once (SEEM) \cite{zou2023segment} to extract \textit{mask embeddings} of all the tangible tokens $\mathcal{V} = {\bv_1, \bv_2, ..., \bv_n}$. These mask embeddings are outputs from an X-Decoder head \cite{zou2022generalized} which encode both localization and object-related information. We set a threshold on the instance segmentation scores to $0.9$ to select only high-scoring detections. We hypothesize that these mask embeddings capture visual information of the corresponding objects within the context of the given image more effectively than extracting separate embeddings for each bounding box using off-the-shelf image encoders like DINO \cite{oquab2024dinov2} or CLIP \cite{radford2021learning}.

In addition to $\mathcal{V}$, we also extract global image features ($\bl$) computed by the segmenter. SEEM, like most segmentation models, follows a U-Net \cite{ronneberger2015unet} design which contracts and expands the image input to result in localized bounding boxes and masks. We therefore collect 2-D average pooled image features at each layer and finally concatenate the resulting vectors.

Next, we extract the set of intangible tokens $\mathcal{U}$ using the Relate-Anything Model (RAM). RAM is built on the Segment-Anything Model (SAM) \cite{alex2023segment} and is trained on the Panoptic Scene Graph Generation dataset (PSG) \cite{yang2022psg} to reason about relationships between any two arbitrary object masks provided with an input image. After extracting the tangible token vectors (mask embeddings) $\mathcal{V}$, we pass pairs of 2D masks corresponding to $(\bv_a, \bv_b), 1 \le a,b \le |\mathcal{V}|$ to RAM to obtain a prediction for the relationship between the corresponding objects. We set a threshold on the classification score at $0.05$ (as specified by the RAM model) to select only high-scoring relationships. We then embed the relationship class using the CLIP text encoder. This process results in a set of intangible tokens $\mathcal{U} = {\bu_1, \bu_2, ..., \bu_m}$.

Finally, we also extract the directional \textit{(subject, object, predicate)} triplet indices, denoted by $\mathcal{E} = \{(a, b, c): 1 \le a, b \le |\mathcal{V}| , 1 \le c \le |\mathcal{U}|, \forall c\}$. This means that the subject $a$ is performing the action $c$, received by the object $b$, and the term \textit{object} in this context refers to the part-of-sentence tag in language terminology ($a,b,c$ are indices of corresponding tokens or visual entities). These triplets capture the semantic correlation between tangible and intangible tokens, resulting in a scene graph of vector nodes and vector edges as illustrated in Figure \ref{fig:seg_rel}. Beyond scene graphs, we also obtain structural information on how objects are co-located in the image by computing the 4-nearest neighbors of each tangible token. Specifically, for every instance we discover from the segmenter, we rank the 4-nearest neighbor instances by computing the Euclidean distance between the centers of the bounding boxes. We formally define this set as $\mathcal{N} = \{(n_k^{(a)}), 1 \le i \le |\mathcal{V}|, \forall a, k \in [1, 2, 3, 4]\}$.

We note that our method of extracting tokens can be substituted with alternative segmentation models or scene graph generation methods \cite{yang2022psg}. Scene graph datasets like Visual Genome \cite{krishna2016visual} provide pre-defined image graphs with object and relationship sets. Our goal of extracting semantic tokens is to capture larger, concrete visual entities as token embeddings rather than tiny patches that are flattened. Therefore, alternate approaches to extract object masks and embeddings can be used in our framework as long as the stated goal is met.

In summary, our token extraction process uses an off-the-shelf segmenter and relation extractor to obtain (i) Image features, denoted by $\bl$, (ii) Set of tangible tokens, denoted by $\mathcal{V} = \{\bv_1, \bv_2, ... \bv_n\}$, (iii) Set of intangible tokens, denoted by $\mathcal{U} = \{\bu_1, \bu_2, ... \bu_m\}$, (iv) Set of (subject, object, predicate) triplets, denoted by $\mathcal{E} = \{(a, b, c): 1 \le a, b \le |\mathcal{V}|, 1 \le c \le |\mathcal{U}|, \forall c\}$, (v) Set of 4-nearest neighbors, denoted by $\mathcal{N} = \{(n_k^{(a)}), 1 \le a \le |\mathcal{V}|, \forall a, k \in [1, 2, 3, 4]\}$. We outline this process in Figure \ref{fig:seg_rel}.

\begin{figure*}
    \centering
    \subcaptionbox*{}{\includegraphics[width=\textwidth,trim={0.5cm 0.5cm 0.3cm 0.4cm},clip]{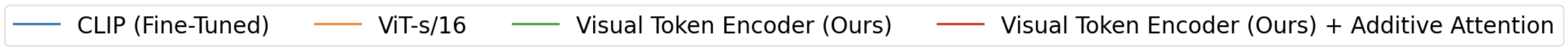}}\\
    \subcaptionbox*{}{\includegraphics[width=0.19\textwidth,trim={0.4cm 0.4cm 0.3cm 0.3cm},clip]{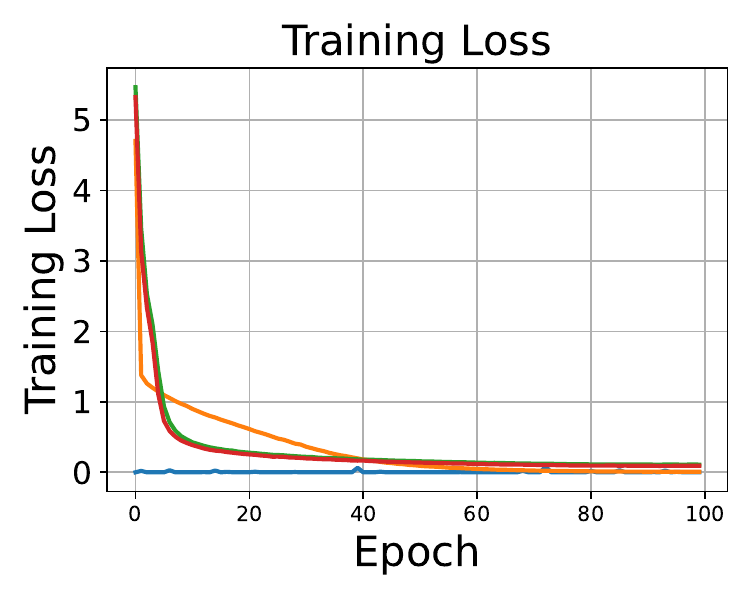}}
    \subcaptionbox*{}{\includegraphics[width=0.19\textwidth,trim={0.4cm 0.4cm 0.3cm 0.3cm},clip]{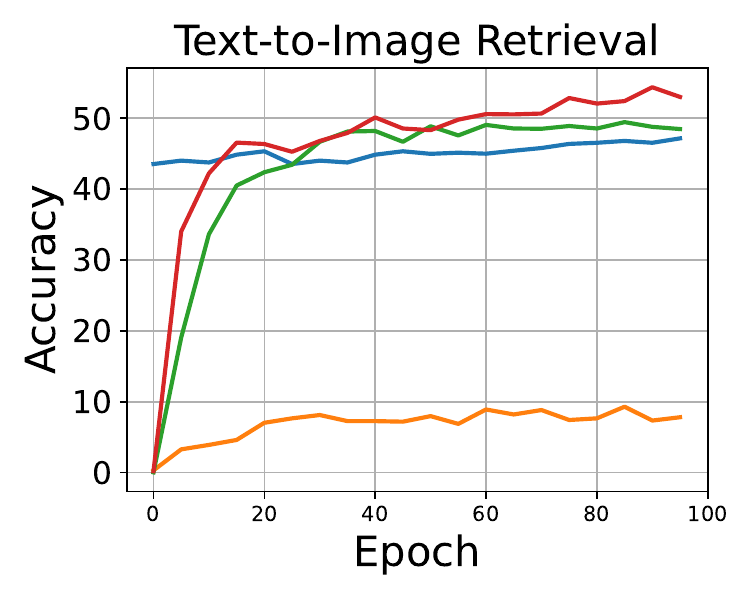}}
    \subcaptionbox*{}{\includegraphics[width=0.19\textwidth,trim={0.4cm 0.4cm 0.3cm 0.3cm},clip]{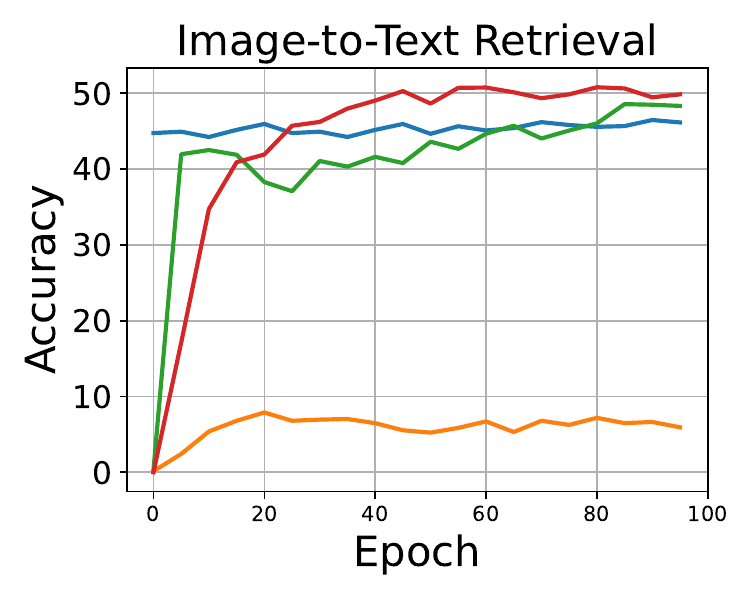}}
    \subcaptionbox*{}{\includegraphics[width=0.19\textwidth,trim={0.4cm 0.4cm 0.3cm 0.3cm},clip]{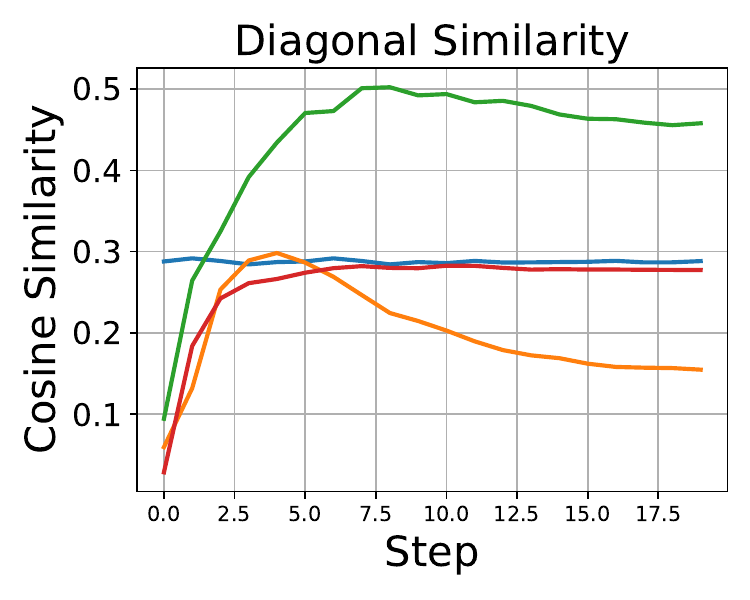}}
    \subcaptionbox*{}{\includegraphics[width=0.19\textwidth,trim={0.4cm 0.4cm 0.3cm 0.3cm},clip]{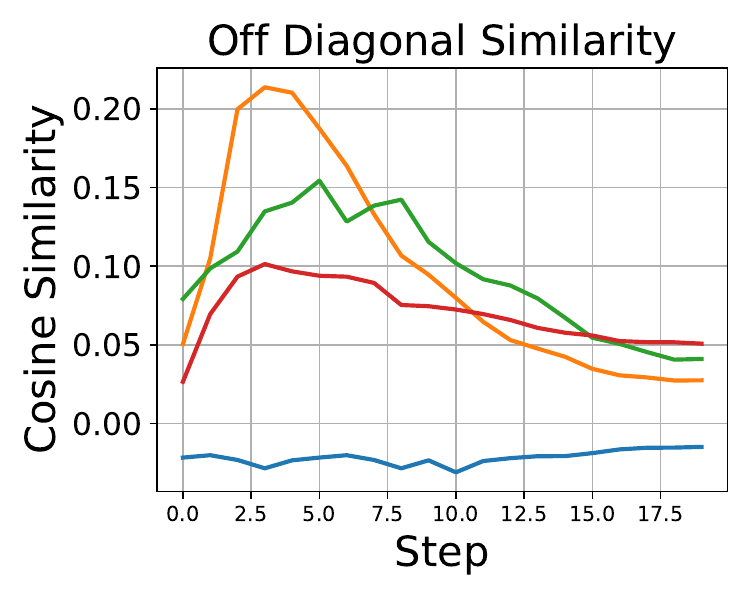}}
    \vspace{-0.5cm}
    \caption{\textbf{Learned Representations:} In text-to-image and image-to-text retrieval accuracy, we observe that our visual token encoders perform best beating both CLIP (fine-tuned) and ViT-s/16 baselines. We also show the average diagonal and off-diagonal similarity of the learned representations across training iterations. From these plots, we observe that the contrast is strongest for our visual token encoder when additive attention is not used. }
    \label{fig:validation}
\end{figure*}
\subsection{Training with Additive Attention}\label{sec:attention}
In this section, we explain the process of training our model using all the elements we extracted in Section \ref{sec:seg_rel}. Rather than training a vision transformer on image data, we train a transformer model $f(\cdot)$ on the extracted token vectors. To this end, we prepare each sample $i$ from the training data as a concatenated set of $\bl_i$ (image features), $\mathcal{V}_i$ (tangible tokens), and $\mathcal{U}_i$ (intangible tokens). We formally define this set as $\mathcal{T}_i = \{\bl_i\} \cup \mathcal{V}_i \cup \mathcal{U}_i$. Similar to text tokens in a transformer, where each token has independent meaning, we now have a set of \textit{visual tokens} where each token has some semantic association. Since each component of $\mathcal{T}_i$ is extracted in a unique manner, involving different deep networks, we add separate positional embeddings to each token based on its type. Specifically, we initialize $3$ learnable positional embeddings $\bp_v$, $\bp_u$, and $\bp_l$, and add them to each token: $\bv = \bv + \bp_v \,  \forall \bv \in \mathcal{V}_i$, $\bu = \bu + \bp_u \, \forall \bu \in \mathcal{U}_i$, $\bl_i = \bl_i + \bp_l$.

The tokens and positional encodings account for most of the visual information available in the given image, including intangible information about actions and relationships. However, the structural connectivity between tokens as illustrated in Figure \ref{fig:seg_rel} is still lacking. Since our data is of graph structure, it makes intuitive sense to use Graph Neural Networks and variants \cite{4700287, kipf2017semisupervised} as primary encoders. However, the scaling and computational complexity of these models make us explore a simpler idea. Transformers, on the other hand, are a proven recipe to train large-scale models efficiently and learn generalizable representations. The attention mechanism \cite{vaswani2023attention} already encapsulates varying levels of importance between tokens and their neighbors.

In the context of text data, attention allows us to identify strong correlations of each word with surrounding words simultaneously as models make sense out of a given sentence. In our setup, we have several tangible tokens, correlated with each other in two manners defined by i) Semantic relations that we extract as intangible tokens $\mathcal{U}$ and (subject, object, predicate) triplets $\mathcal{E}$ and ii) Relative positions in the image defined by the set of nearest neighbors $\mathcal{N}$. We therefore simulate both of these correlations by applying a ranked additive weight to the computed attention scores between each pair of tokens.

We prepare a weight matrix $A_i \in \bbR^{|\mathcal{T}| \times |\mathcal{T}|}$ for any $i^{th}$ data sample. We populate $A_i$ with $7$ types of relationships between the tokens in $\mathcal{T}_i$, ranked by their importance in image comprehension.

\begin{align}
    A_i^{(a, b)} = \begin{cases}
    7, & \text{if } (a, b, .) \in \mathcal{E}_i \\
    6, & \text{if } (a, ., b) \in \mathcal{E}_i \text{ or } (., a, b) \in \mathcal{E}_i \\
    5, & \text{if } (b, ., a) \in \mathcal{E}_i \text{ or } (., b, a) \in \mathcal{E}_i \\
    5 - k, & \text{if } b = \mathcal{N}_i^{(k)}, k = \{1,2,3,4\} \\
\end{cases}
\end{align}

Here, $a$ and $b$ represent the indices of any pair of tokens in the given set $\mathcal{T}_i$ and for simplicity of notation, we directly denote (subject, object, predicate) sets using $a$ and $b$. With reference to the scene graph shown in Figure \ref{sec:seg_rel}, the ranks are applied as (i) $7$ for a directional node to node connection, (ii) $6$ for every node to edge connection (\textit{person} to \textit{beside}, \textit{tree} to \textit{beside}), (iii) $5$ for edge to node connections (\textit{beside} to \textit{person} and \textit{beside} to \textit{tree}), (iv) $4$, $3$, $2$, $1$ for the first to the fourth nearest neighbor (in the image) of the given node respectively. 

We define an attention weight encoding $\ba \in \bbR^8$ that is learned based on the rankings in the attention weight matrix $A_i$ for all samples. The first element in $\ba$ is set to $0$ and each subsequent element is learned such that the difference between that element and the previous element is equal to a rank $\{1, 2, .. 7\}$. This is done simply by computing the cumulative sum, $cumsum(\exp(\ba))$ and substituting each rank in $A_i$ with the learned rank weight in $cumsum(\exp(\ba))$. Finally, we add this updated learned weight matrix $A_i$ to the computed self-attention score across all attention heads in our model. 

We deploy our \textit{visual token encoder} that applies all of the strategies described above in a vision-language pre-training setup. As shown in Figure \ref{fig:framework}, we preprocess every image in the training data using our visual tokenization method, which extracts the set of tokens $\mathcal{T}$ and prepares attention rank matrices $A$ for those tokens. The token embeddings are padded to a fixed context length and added to positional encodings $\bp_v$, $\bp_u$, and $\bp_l$. The tokens, along with the attention weight matrix (computed using $\ba$ and $A_i$), are fed into our visual token encoder, $f(\cdot)$, which follows a standard transformer architecture to extract fixed-length image embeddings $\bs$. We simultaneously also train a text encoder, $g(\cdot)$ (also a transformer), on the image captions to extract fixed-length text embeddings $\bt$. We follow the CLIP \cite{radford2021learning} optimization, which applies a simple contrastive loss between all $s_i$'s and $t_i$'s in large batches.

Practically, training this model is more efficient compared to ViTs and CLIP since we process relatively low-dimensional data compared to high-dimensional large images. We use a context length of $77$ during token extraction and our token embedding width is $512$ (arising from the segmentation model). Therefore, each sample is of $77\times512$ dimensions along with $A$, a $|\mathcal{T}|\times|\mathcal{T}|$ dimensional weight matrix, where $|\mathcal{T}| < 77$. While the training speed and compute cost is significantly lower, we cannot ignore the added overhead of token extraction itself. The compute cost and memory overhead comes from the segmentation and relation extraction process where each image needs to be processed individually and the extracted token and metadata need to be saved on the disk for training.
\begin{table*}
    \centering
    \caption{\textbf{ARO Benchmark:} We evaluate our model and baselines on $4$ components of the ARO \cite{yuksekgonul2023visionlanguage} benchmark and measure the accuracy for each. We outperform both CLIP and ViT across VG-Relation, VG-Attribution and COCO-Order and beats the ViT on Flickr-Order. * indicates models that are trained from scratch on COCO. FT indicates pretrained models that are fine-tuned on COCO.}
    \resizebox{\textwidth}{!}{
    \begin{tabular}{c|c|c|c|c|c}
    \toprule
    \textbf{Image Encoder} & \textbf{Text Encoder} & \textbf{VG-Relation} & \textbf{VG-Attribution} & \textbf{COCO-Order} & \textbf{Flickr-Order} \\
    \midrule
    \midrule
    CLIP & CLIP &  59.9 & 63.1 & 47.4 & \textbf{58.0} \\
    CLIP\textsuperscript{FT} & CLIP\textsuperscript{FT} & 65.8 & 65.9 & 56.4 & 32.7 \\
    \midrule
    ViT-s/16* & CLIP\textsuperscript{FT} & 53.5 & 53.9 & 38.7 & 28.9 \\
    Visual Token Encoder* (Ours) & \multirow{2}{*}{CLIP\textsuperscript{FT}} & \multirow{2}{*}{67.8} & \multirow{2}{*}{64.1} & \multirow{2}{*}{17.2} & \multirow{2}{*}{34.8} \\
     (without additive attn.) & & & & &\\
    Visual Token Encoder* (Ours) & CLIP\textsuperscript{FT} & \textbf{68.9} & \textbf{66.2} & \textbf{56.8} & 41.4\\
    \bottomrule
    \end{tabular}
    }
    \label{tab:aro}
\end{table*}
\begin{table*}
    \centering
    \caption{\textbf{Winoground Benchmark:} We evaluate our model and baselines on Winoground \cite{Thrush_2022} and measure the $3$ metrics given by the dataset. Our model outperforms CLIP and ViT in the image correct and group correct metrics. * indicates models that are trained from scratch on COCO. FT indicates pretrained models that are fine-tuned on COCO.}
    \resizebox{0.8\textwidth}{!}{
    \begin{tabular}{c|c|c|c|c}
    \toprule
    \textbf{Image Encoder} & \textbf{Text Encoder} & \textbf{Text Correct} & \textbf{Image Correct} & \textbf{Group Correct}  \\
    \midrule
    \midrule
    CLIP & CLIP & 30.75 & 10.50 & 8.00 \\
    CLIP\textsuperscript{FT} & CLIP\textsuperscript{FT} & 28.25 & 12.00 & 7.25 \\
    \midrule
    ViT-s/16* & CLIP\textsuperscript{FT} & 18.00 & 13.00 & 7.00 \\
    Visual Token Encoder* (Ours) & \multirow{2}{*}{CLIP\textsuperscript{FT}} & \multirow{2}{*}{\textbf{28.25}} & \multirow{2}{*}{15.25} & \multirow{2}{*}{9.25} \\
     (without additive attn.) & & & & \\
    Visual Token Encoder* (Ours) & CLIP\textsuperscript{FT} & 27.00 & \textbf{16.00} & \textbf{9.75} \\
    \bottomrule
    \end{tabular}
}
    \label{tab:winoground}
\end{table*}

\section{Results}\label{sec:results}
\subsection{Experimental Setup}\label{sec:exp_setup}
Our experimental premise is to demonstrate a proof-of-concept of our hypothesis stating - using semantically meaningful tokens can be beneficial in learning comprehensive, compositional representations. For token extraction, as discussed in Section \ref{sec:seg_rel}, we use the segmenter, SEEM \cite{alex2023segment} (Focal-L \cite{yang2022focal} backbone) and relation extractor RAM \cite{yang2022psg}. SEEM is trained on COCO \cite{lin2015microsoft} while RAM is trained on the Panoptic Scene Graph Generation (PSG) dataset which contains $49$K images arising from COCO and Visual Genome \cite{krishna2016visual}. We re-train RAM to use the segmentation outputs from SEEM rather than SAM \cite{alex2023segment} to maintain consistency. We extract and save all sets of tokens and metadata as listed in Section \ref{sec:seg_rel} for the COCO train ($118$M samples) and validation ($5$K samples) splits. We set our context length to $77$ tokens and add zero-padding, if needed. 

Next, we train our Visual Token Encoder from scratch on the synthesized COCO token dataset to confirm our hypothesis. Our tangible and intangible tokens are of $512$ dimensions and we use a $3$-layer MLP with ReLU \cite{agarap2018deep} activation to project the concatenated intermediate image features from the segmentation model backbone into $512$ dimensions. We use the PyTorch implementation of the Transformer model using $8$ layers and $8$ attention heads with a linear projection head that dimension of $512$. We simultaneously fine-tune the CLIP (ViT-B/32) text encoder, which outputs $512$-dimensional text embeddings, on the COCO caption data by loading $1$ randomly sampled caption out of $5$ per image in the dataset.  

We sweep over $4$ learning rates $\{1e^{-6}, 5e^{-6}, 1e^{-5}, 5e^{-5}\}$ and choose the best performing model. We use the AdamW optimizer \cite{loshchilov2019decoupled} and train for $100$ epochs, using a batch size of $256$, with a warmup of $10$ epochs and cosine annealing learning rate schedule. We perform experiments with and without additive attention as an ablation.

In order to understand the benefits of using our proposed tokenization approach, we compare with $3$ baseline setups which are directly trained on image data using standard tokenization techniques i.e., image patchification. The first setup replaces our tokenizer and transformer with a standard ViT \cite{dosovitskiy2021image} trained directly on COCO images. We use the PyTorch implementation of the VisionTransformer model and train a ViT-s/16 variant which has $8$ layers and $8$ attention heads, closely matching the architecture of our Visual Token Encoder. We align the learned image embeddings with the CLIP text embeddings in the same manner as described above. Our second setup, is the pre-trained CLIP (ViT-B/32) model which is trained on very large-scale data \cite{radford2021learning} of roughly $400$M samples. Finally, in our last setup, we fine-tune the pre-trained CLIP (ViT-B/32) model on COCO images and captions. This setup is similar to that of the ViT-s/16 except CLIP is already pre-trained on a large amount of data while the ViT is trained from scratch. We use the same training pipeline described for our Visual Token Encoder for all experiments - including number of epochs, learning rate sweeps, optimizer, schedulers, etc. and all results are averaged over $2$ random seeds. Our Visual Token Encoder can be trained efficiently using $2$ A$5000$'s, however, the ViT-s/16 and CLIP models need to be trained on $4$ A$6000$'s.

\subsection{Learned Representations}
We measure several metrics through the course of training our model and baselines to understand the quality of the learned visual representations. As mentioned, in our training pipeline, we fine-tune the CLIP text encoder with a vision component which can be any one of (i) Visual Token Encoder (Ours), (ii) Visual Token Encoder (Ours), without Additive Attention, (iii) ViT-s/16 (iv) CLIP (fine-tuned). In Figure \ref{fig:validation}, we show that the training loss converges across all setups. The CLIP model which is already pre-trained maintains a low loss through the course of training.

We evaluate the alignment between visual and text representations by calculating the image-to-text and text-to-image retrieval scores on the COCO validation split across iterations. We randomly sample $1$ caption out of $5$ per COCO sample and compute all text and visual embeddings. We then measure the zero-shot retrieval accuracy of matching the closest image to a given text (text-to-image) and closest text to a given image (image-to-text). These metrics are plotted in the second and third subfigures in Figure \ref{fig:validation}. We observe that the retrieval scores of our Visual Token Encoder are the highest amongst all experiments. The usage of additive attention results in $54.35$ text-to-image retrieval accuracy which is a $9\%$ improvement from the fine-tuned CLIP model and $47\%$ improvement over ViT-s/16. This model also performs best for image-to-text retrieval achieving $49.76$ accuracy which is $4\%$ improvement over the fine-tuned CLIP model and a $44\%$ improvement over ViT-s/16. These results are especially noteworthy because both our Visual Token Encoder and the ViT-s/16 are trained from scratch to convergence on the same data (COCO) but our method learns significantly more powerful representations. Our model also beats CLIP which is pre-trained on $1000\times$ more data and fine-tuned on COCO for the same number of iterations. 

In the last two plots in Figure \ref{fig:validation}, we compare the average diagonal and off-diagonal similarities in the COCO validation text-image embedding cosine similarity matrix. Since the CLIP model has already converged, we do not observe major changes in embedding similarity as training progresses. Other models which are trained from scratch show an increasing trajectory in the off-diagonal similarity, followed by a decrease, finally leading to a similarity score lower than that of the diagonal values. We observe the strongest contrast between diagonal and off-diagonal scores in our visual token encoder when additive attention is not used and the weakest contrast in the ViT-s/16 model. 

\subsection{Compositionality Benchmarks}
As vision-language models gained popularity, several follow up works have challenged their compositional reasoning capabilities. Compositionality benchmarks like ARO \cite{yuksekgonul2023visionlanguage} and Winoground \cite{Thrush_2022} propose a set of evaluation datasets which can be used to understand the depth of vision-language model reasoning. Both benchmarks have highlighted the significant lack of compositional understanding in state-of-the-art vision language models like CLIP. In this section, we study the behavior of the Visual Token Encoder we proposed on these benchmarks, compared to our baselines.

The ARO benchmark consists of $4$ datasets - Visual Genome-Relation (VG-Relation), Visual Genome-Attribution (VG-Attribution), COCO-Order and Flickr-Order. A sample arising from VG-Relation and VG-Attribution consists of an image with $2$ caption options, a correct caption and an incorrect caption where either the relations (between objects) or the attributions (object properties) are interchanged across objects. Samples from COCO-Order and Flickr-Order consist of an image and $5$ caption options, where only one is correct and the others have shuffled words to test order sensitivity. The accuracy for each dataset measures the percentage of images matched with the correct caption by the given model using its corresponding similarity metric (cosine similarity). We evaluate our model and baselines on each of these datasets and present our results in Table \ref{tab:aro}.

In VG-Relation, VG-Attribution and COCO-Order benchmarks, our model with additive attention performs best, showing a $10\%$ improvement over CLIP (off-the-shelf) and $18\%$ over ViT-s/16. We consider the ViT-s/16 as a fair competitor with our model since it has seen the same data (COCO train) and Visual Genome and Flickr are both out-of-distribution datasets. In COCO-Order, we observe a degraded accuracy in our model when additive attention is not used. Without additive attention, the visual tokens of COCO are simply stacked and presented to the transformer with no information of the nature of their relations. We suspect that this prevents the model from choosing the correct permutation of words in COCO-Order. Our model outperforms the ViT in Flickr-Order but does not beat the CLIP (off-the-shelf) baseline. This may be because Flickr may be closer to the training distribution of $400M$ samples that CLIP has seen. 

Winoground, like ARO, also tests for relation and attribution reasoning. Each sample in Winoground consists of $2$ images and $2$ captions and the accuracy is measured by a given model's capability of associating the correct image to the correct caption and vice versa. The resulting metrics are Text Correct (assigning text to correct image), Image Correct (assigning image to correct text) and Group Correct (a combination of the previous two). In Table \ref{tab:winoground}, we summarize these metrics across our experiments. Compared to ARO, Winoground is a harder benchmark where even a large-scale model like CLIP only reaches a $10.50$ image correct score. Our model outperforms all others in image correct and group correct scores showing $3\%$ improvement over ViT and $4\%$ over CLIP (fine-tuned). We beat the ViT by $10\%$ on text correct scores with and without additive attention. Our tokenization process is beneficial to the image correct scores rather than text correct, because we attempt to learn compositional image embeddings such that they are better associated with correct captions. 
\section{Discussion}
We challenge the premise of equal-sized patching in vision transformers and propose to use variable-sized, semantically meaningful tokens for visual understanding. We use off-the-shelf segmentation models and scene graph generation models which can detect high-level patches, such as objects in real images, which possess independent physical meaning. Additionally, we show that visual comprehension can be enhanced with intangible tokens like actions and relations that have semantic significance but are not physically localized in the image. We train a transformer model, referred to as the Visual Token Encoder, on the extracted set of tokens on the COCO dataset to learn image representations and align them with caption representations from a fine-tuned CLIP text encoder. We incorporate other metadata, such as directional relationships and relative positions of tokens, by applying additive attention weights ranked by importance. These updates result in a $47\%$ improvement in text-to-image retrieval compared to using a Vision Transformer (ViT) and a $9\%$ improvement over the fine-tuned CLIP model. Additionally, we show an $18\%$ improvement over ViT in the ARO benchmark and a $10\%$ improvement in the Winoground benchmark, indicating that our Visual Token Encoder produces higher-quality compositional representations. Our contribution presents a proof-of-concept for rethinking tokenization in vision models and the associated potential benefits. Our findings open new avenues for empirical and theoretical research, specifically: (i) How does this tokenization approach perform in large-scale setups? (ii) Can we develop a unified model for both scene-graph generation and representation learning? (iii) How can our tokenization method be made more compute and memory efficient?
\section{Acknowledgement}
This project was supported in part by an award from Capital One, a grant from an NSF CAREER AWARD 1942230, ONR YIP award N00014-22-1-2271, ARO’s Early Career Program Award 310902-00001, HR00112090132 (DARPA/RED), HR001119S0026 (DARPA/GARD), Army Grant No. W911NF2120076, the NSF award CCF2212458, NSF Award No. 2229885 (NSF Institute for Trustworthy AI in Law and Society, TRAILS), an Amazon Research Award.

{
    \small
    \bibliographystyle{ieeenat_fullname}
    \bibliography{bibliography}

\begin{thebibliography}{33}
\providecommand{\natexlab}[1]{#1}
\providecommand{\url}[1]{\texttt{#1}}
\expandafter\ifx\csname urlstyle\endcsname\relax
  \providecommand{\doi}[1]{doi: #1}\else
  \providecommand{\doi}{doi: \begingroup \urlstyle{rm}\Url}\fi

\bibitem[Agarap(2018)]{agarap2018deep}
Abien~Fred Agarap.
\newblock Deep learning using rectified linear units (relu), 2018.

\bibitem[Chen et~al.(2021)Chen, Zhu, Zhao, Hu, Zeng, Wang, and Tang]{Chen_2021}
Zhiyang Chen, Yousong Zhu, Chaoyang Zhao, Guosheng Hu, Wei Zeng, Jinqiao Wang, and Ming Tang.
\newblock Dpt: Deformable patch-based transformer for visual recognition.
\newblock In \emph{Proceedings of the 29th ACM International Conference on Multimedia}. ACM, 2021.

\bibitem[Deng et~al.(2009)Deng, Dong, Socher, Li, Li, and Fei-Fei]{imagenet}
Jia Deng, Wei Dong, Richard Socher, Li-Jia Li, Kai Li, and Li Fei-Fei.
\newblock Imagenet: A large-scale hierarchical image database.
\newblock In \emph{2009 IEEE Conference on Computer Vision and Pattern Recognition}, pages 248--255, 2009.

\bibitem[Dosovitskiy et~al.(2021)Dosovitskiy, Beyer, Kolesnikov, Weissenborn, Zhai, Unterthiner, Dehghani, Minderer, Heigold, Gelly, Uszkoreit, and Houlsby]{dosovitskiy2021image}
Alexey Dosovitskiy, Lucas Beyer, Alexander Kolesnikov, Dirk Weissenborn, Xiaohua Zhai, Thomas Unterthiner, Mostafa Dehghani, Matthias Minderer, Georg Heigold, Sylvain Gelly, Jakob Uszkoreit, and Neil Houlsby.
\newblock An image is worth 16x16 words: Transformers for image recognition at scale, 2021.

\bibitem[Han et~al.(2022)Han, Wang, Guo, Tang, and Wu]{han2022vision}
Kai Han, Yunhe Wang, Jianyuan Guo, Yehui Tang, and Enhua Wu.
\newblock Vision gnn: An image is worth graph of nodes, 2022.

\bibitem[He et~al.(2018)He, Gkioxari, Dollár, and Girshick]{he2018mask}
Kaiming He, Georgia Gkioxari, Piotr Dollár, and Ross Girshick.
\newblock Mask r-cnn, 2018.

\bibitem[Kipf and Welling(2017)]{kipf2017semisupervised}
Thomas~N. Kipf and Max Welling.
\newblock Semi-supervised classification with graph convolutional networks, 2017.

\bibitem[Kirillov et~al.(2019)Kirillov, He, Girshick, Rother, and Dollár]{kirillov2019panoptic}
Alexander Kirillov, Kaiming He, Ross Girshick, Carsten Rother, and Piotr Dollár.
\newblock Panoptic segmentation, 2019.

\bibitem[Kirillov et~al.(2023)Kirillov, Mintun, Ravi, Mao, Rolland, Gustafson, Xiao, Whitehead, Berg, Lo, Dollár, and Girshick]{alex2023segment}
Alexander Kirillov, Eric Mintun, Nikhila Ravi, Hanzi Mao, Chloe Rolland, Laura Gustafson, Tete Xiao, Spencer Whitehead, Alexander~C. Berg, Wan-Yen Lo, Piotr Dollár, and Ross Girshick.
\newblock Segment anything, 2023.

\bibitem[Krishna et~al.(2016)Krishna, Zhu, Groth, Johnson, Hata, Kravitz, Chen, Kalantidis, Li, Shamma, Bernstein, and Li]{krishna2016visual}
Ranjay Krishna, Yuke Zhu, Oliver Groth, Justin Johnson, Kenji Hata, Joshua Kravitz, Stephanie Chen, Yannis Kalantidis, Li-Jia Li, David~A. Shamma, Michael~S. Bernstein, and Fei-Fei Li.
\newblock Visual genome: Connecting language and vision using crowdsourced dense image annotations, 2016.

\bibitem[Li et~al.(2022)Li, Li, Xiong, and Hoi]{li2022blip}
Junnan Li, Dongxu Li, Caiming Xiong, and Steven Hoi.
\newblock Blip: Bootstrapping language-image pre-training for unified vision-language understanding and generation, 2022.

\bibitem[Li et~al.(2019)Li, Yatskar, Yin, Hsieh, and Chang]{li2019visualbert}
Liunian~Harold Li, Mark Yatskar, Da Yin, Cho-Jui Hsieh, and Kai-Wei Chang.
\newblock Visualbert: A simple and performant baseline for vision and language, 2019.

\bibitem[Lin et~al.(2015)Lin, Maire, Belongie, Bourdev, Girshick, Hays, Perona, Ramanan, Zitnick, and Dollár]{lin2015microsoft}
Tsung-Yi Lin, Michael Maire, Serge Belongie, Lubomir Bourdev, Ross Girshick, James Hays, Pietro Perona, Deva Ramanan, C.~Lawrence Zitnick, and Piotr Dollár.
\newblock Microsoft coco: Common objects in context, 2015.

\bibitem[Liu et~al.(2021)Liu, Lin, Cao, Hu, Wei, Zhang, Lin, and Guo]{liu2021swin}
Ze Liu, Yutong Lin, Yue Cao, Han Hu, Yixuan Wei, Zheng Zhang, Stephen Lin, and Baining Guo.
\newblock Swin transformer: Hierarchical vision transformer using shifted windows, 2021.

\bibitem[Long et~al.(2015)Long, Shelhamer, and Darrell]{long2015fully}
Jonathan Long, Evan Shelhamer, and Trevor Darrell.
\newblock Fully convolutional networks for semantic segmentation, 2015.

\bibitem[Loshchilov and Hutter(2019)]{loshchilov2019decoupled}
Ilya Loshchilov and Frank Hutter.
\newblock Decoupled weight decay regularization, 2019.

\bibitem[Ma et~al.(2024)Ma, Jiang, Wu, Yuan, and Qi]{ma2024groma}
Chuofan Ma, Yi Jiang, Jiannan Wu, Zehuan Yuan, and Xiaojuan Qi.
\newblock Groma: Localized visual tokenization for grounding multimodal large language models, 2024.

\bibitem[Oquab et~al.(2024)Oquab, Darcet, Moutakanni, Vo, Szafraniec, Khalidov, Fernandez, Haziza, Massa, El-Nouby, Assran, Ballas, Galuba, Howes, Huang, Li, Misra, Rabbat, Sharma, Synnaeve, Xu, Jegou, Mairal, Labatut, Joulin, and Bojanowski]{oquab2024dinov2}
Maxime Oquab, Timothée Darcet, Théo Moutakanni, Huy Vo, Marc Szafraniec, Vasil Khalidov, Pierre Fernandez, Daniel Haziza, Francisco Massa, Alaaeldin El-Nouby, Mahmoud Assran, Nicolas Ballas, Wojciech Galuba, Russell Howes, Po-Yao Huang, Shang-Wen Li, Ishan Misra, Michael Rabbat, Vasu Sharma, Gabriel Synnaeve, Hu Xu, Hervé Jegou, Julien Mairal, Patrick Labatut, Armand Joulin, and Piotr Bojanowski.
\newblock Dinov2: Learning robust visual features without supervision, 2024.

\bibitem[Radford et~al.(2021)Radford, Kim, Hallacy, Ramesh, Goh, Agarwal, Sastry, Askell, Mishkin, Clark, Krueger, and Sutskever]{radford2021learning}
Alec Radford, Jong~Wook Kim, Chris Hallacy, Aditya Ramesh, Gabriel Goh, Sandhini Agarwal, Girish Sastry, Amanda Askell, Pamela Mishkin, Jack Clark, Gretchen Krueger, and Ilya Sutskever.
\newblock Learning transferable visual models from natural language supervision, 2021.

\bibitem[Ronneberger et~al.(2015)Ronneberger, Fischer, and Brox]{ronneberger2015unet}
Olaf Ronneberger, Philipp Fischer, and Thomas Brox.
\newblock U-net: Convolutional networks for biomedical image segmentation, 2015.

\bibitem[Scarselli et~al.(2009)Scarselli, Gori, Tsoi, Hagenbuchner, and Monfardini]{4700287}
Franco Scarselli, Marco Gori, Ah~Chung Tsoi, Markus Hagenbuchner, and Gabriele Monfardini.
\newblock The graph neural network model.
\newblock \emph{IEEE Transactions on Neural Networks}, 20\penalty0 (1):\penalty0 61--80, 2009.

\bibitem[Singh et~al.(2022)Singh, Hu, Goswami, Couairon, Galuba, Rohrbach, and Kiela]{Singh_2022}
Amanpreet Singh, Ronghang Hu, Vedanuj Goswami, Guillaume Couairon, Wojciech Galuba, Marcus Rohrbach, and Douwe Kiela.
\newblock Flava: A foundational language and vision alignment model.
\newblock In \emph{2022 IEEE/CVF Conference on Computer Vision and Pattern Recognition (CVPR)}. IEEE, 2022.

\bibitem[Thrush et~al.(2022)Thrush, Jiang, Bartolo, Singh, Williams, Kiela, and Ross]{Thrush_2022}
Tristan Thrush, Ryan Jiang, Max Bartolo, Amanpreet Singh, Adina Williams, Douwe Kiela, and Candace Ross.
\newblock Winoground: Probing vision and language models for visio-linguistic compositionality.
\newblock In \emph{2022 IEEE/CVF Conference on Computer Vision and Pattern Recognition (CVPR)}. IEEE, 2022.

\bibitem[Touvron et~al.(2020)Touvron, Cord, Douze, Massa, Sablayrolles, and Jégou]{touvron2020training}
Hugo Touvron, Matthieu Cord, Matthijs Douze, Francisco Massa, Alexandre Sablayrolles, and Hervé Jégou.
\newblock Training data-efficient image transformers and distillation through attention, 2020.

\bibitem[Vaswani et~al.(2023)Vaswani, Shazeer, Parmar, Uszkoreit, Jones, Gomez, Kaiser, and Polosukhin]{vaswani2023attention}
Ashish Vaswani, Noam Shazeer, Niki Parmar, Jakob Uszkoreit, Llion Jones, Aidan~N. Gomez, Lukasz Kaiser, and Illia Polosukhin.
\newblock Attention is all you need, 2023.

\bibitem[Xia et~al.(2023)Xia, Pan, Song, Li, and Huang]{xia2023dat}
Zhuofan Xia, Xuran Pan, Shiji Song, Li~Erran Li, and Gao Huang.
\newblock Dat++: Spatially dynamic vision transformer with deformable attention, 2023.

\bibitem[Yang et~al.(2022{\natexlab{a}})Yang, Ang, Guo, Zhou, Zhang, and Liu]{yang2022psg}
Jingkang Yang, Yi~Zhe Ang, Zujin Guo, Kaiyang Zhou, Wayne Zhang, and Ziwei Liu.
\newblock Panoptic scene graph generation.
\newblock In \emph{ECCV}, 2022{\natexlab{a}}.

\bibitem[Yang et~al.(2022{\natexlab{b}})Yang, Li, Dai, Yuan, and Gao]{yang2022focal}
Jianwei Yang, Chunyuan Li, Xiyang Dai, Lu Yuan, and Jianfeng Gao.
\newblock Focal modulation networks, 2022{\natexlab{b}}.

\bibitem[Yuksekgonul et~al.(2023)Yuksekgonul, Bianchi, Kalluri, Jurafsky, and Zou]{yuksekgonul2023visionlanguage}
Mert Yuksekgonul, Federico Bianchi, Pratyusha Kalluri, Dan Jurafsky, and James Zou.
\newblock When and why vision-language models behave like bags-of-words, and what to do about it?, 2023.

\bibitem[Zeng et~al.(2022)Zeng, Zhang, and Li]{zeng2022multigrained}
Yan Zeng, Xinsong Zhang, and Hang Li.
\newblock Multi-grained vision language pre-training: Aligning texts with visual concepts, 2022.

\bibitem[Zhou et~al.(2018)Zhou, Zhao, Puig, Xiao, Fidler, Barriuso, and Torralba]{zhou2018semantic}
Bolei Zhou, Hang Zhao, Xavier Puig, Tete Xiao, Sanja Fidler, Adela Barriuso, and Antonio Torralba.
\newblock Semantic understanding of scenes through the ade20k dataset, 2018.

\bibitem[Zou et~al.(2022)Zou, Dou, Yang, Gan, Li, Li, Dai, Behl, Wang, Yuan, Peng, Wang, Lee, and Gao]{zou2022generalized}
Xueyan Zou, Zi-Yi Dou, Jianwei Yang, Zhe Gan, Linjie Li, Chunyuan Li, Xiyang Dai, Harkirat Behl, Jianfeng Wang, Lu Yuan, Nanyun Peng, Lijuan Wang, Yong~Jae Lee, and Jianfeng Gao.
\newblock Generalized decoding for pixel, image, and language, 2022.

\bibitem[Zou et~al.(2023)Zou, Yang, Zhang, Li, Li, Wang, Wang, Gao, and Lee]{zou2023segment}
Xueyan Zou, Jianwei Yang, Hao Zhang, Feng Li, Linjie Li, Jianfeng Wang, Lijuan Wang, Jianfeng Gao, and Yong~Jae Lee.
\newblock Segment everything everywhere all at once, 2023.

\end{thebibliography}
}


\end{document}